\documentclass[letterpaper, 10 pt, conference]{ieeeconf}  

\IEEEoverridecommandlockouts                              

\usepackage{graphicx}
\usepackage[ruled,vlined]{algorithm2e}
\usepackage{mathtools}

\usepackage{float} 
\usepackage{amsmath} 

\DeclareMathOperator*{\argmin}{arg\,min}

\usepackage{bbold}
\usepackage{bbm}

\usepackage{textcomp}

\title{\LARGE \bf
Affordance Template Registration via Human-in-the-loop Corrections
}

\author{Michael Hagenow$^{1}$, Michael Zinn$^{1}$, Terrence Fong$^{2}$, Evan Laske$^{3}$, and Kimberly Hambuchen$^{3}$
\thanks{This work was supported in part by a NASA University Leadership Initiative (ULI) grant awarded to the UW-Madison and The Boeing Company (Cooperative Agreement \# 80NSSC19M0124).}
\thanks{$^{1}$Michael Hagenow and Michael Zinn are with the Department of Mechanical
Engineering, University of Wisconsin--Madison, Madison 53706, USA
        {\tt\small [mhagenow|mzinn]@wisc.edu}}%
\thanks{$^{2}$Terrence Fong is with the Intelligent Robotics Group, NASA Ames Research Center, Mountain View, CA 94035, USA {\tt\small terrence.w.fong@nasa.gov}}%
\thanks{$^{3}$Evan Laske and Kimberly Hambuchen are with NASA Johnson Space Center, 2101 NASA Pkwy, Houston, TX 77058, USA {\tt\small [evan.laske|kimberly.a.hambuchen]@nasa.gov}}%
}

\begin{document}

\maketitle
\thispagestyle{empty}
\pagestyle{empty}

\begin{abstract}
Affordance Templates (ATs) are a method for parameterizing objects for autonomous robot manipulations. In this approach, instances of an object are registered by positioning a model in a 3D environment, which requires a large amount of user input. We instead propose a registration method which combines autonomy and user corrections. For selected objects, the system determines both the model and corresponding pose autonomously. The user makes corrections only if the model or pose is incorrect. This method increases the level of autonomy compared to existing approaches which can reduce user input and time on task. In this paper, we present an overview of existing methods, a description of our method, preliminary results, and planned future work.
\end{abstract}
\section{INTRODUCTION}
Many applications require the non-collocated use of robotics including space operations, disaster recovery, and improvised explosive device (IED) response. To enable robots to perform tasks asynchronously in these environments, objects of interest need to be both localized and parameterized. One way to specify important objects and corresponding object information is through Affordance Templates (ATs) \cite{hart2014affordance}\cite{HartAffordance2015}. In this paradigm, users create object templates which specify relevant information, such as the geometry, joint limits, torque limits, and permissible states (e.g., open/closed). Once the template is created, an operator can identify instances of an object in a given 3D environment by positioning a proxy mesh via interactive object markers. While Affordance Templates are a powerful standard for planning object manipulations, the preliminary step of iteratively and carefully positioning the interactive markers in the environment leads to a large amount of user input, which as a consequence increases the time spent on tasks. We instead propose a method to minimize the user input during object registration by combining an automated fitting system with intermittent user corrections.

\begin{figure}[]
\centering
\includegraphics[width=3.30in]{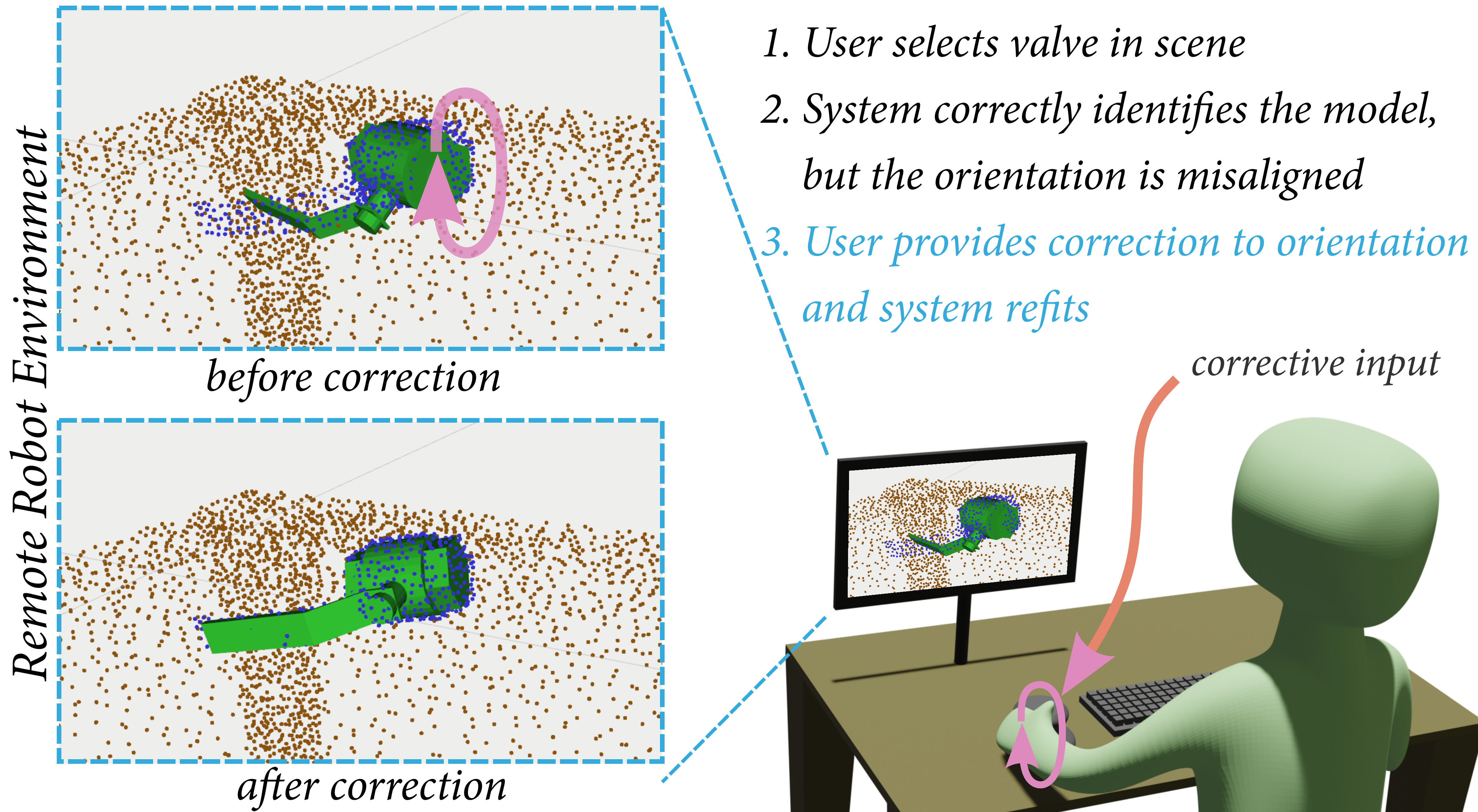}
\vspace{-5pt}
\caption{An operator selects a point in a remote robot environment using a screen-based interface. \emph{Top left}: the system automatically fits the model, but the orientation converges to a local minimum. \emph{Bottom left}: using a 6D input, the user provides a small correction to the orientation from which the system is able to refit to the correct object pose.}
\label{fig:teaser}
\vspace{-15pt}
\end{figure}

Previous methods have leveraged human-in-the-loop techniques for object manipulation, but have not focused on the use of corrections. Masnadi et al. \cite{masnadi2019sketching}\cite{masnadi2020affordit} propose methods for affordance specification by sketching relevant articulations on a touchscreen interface. Kent et al. \cite{kent2020leveraging} propose a method for human-in-the-loop object grasp specification based on user selection from a small list of candidate grasp poses. Jorgensen et al. \cite{jorgensen2019deploying} use Iterative Closest Points (ICP) to automatically register meshes to rough positioning of object models. Finally, Ye et al. \cite{ye2021human} propose a snap-to-grid approach using Monte Carlo localization where objects are first roughly positioned using interactive markers, followed by automated system pose refinement. In this work, we propose to further increase the system level of autonomy by using corrections as input \cite{hagenow2021informing}\cite{HagenowCSA2021}. As exemplified in Figure \ref{fig:teaser}, the system can identify models and pose automatically and the user only intervenes when the system is incorrect, which can further reduce the required user input compared to other human-in-the-loop techniques. We believe this approach has the potential to reduce the time taken to register ATs.

\section{APPROACH}
\subsection{Technical}
\label{sec:tech}
To enable semi-automated fitting of Affordance Templates, we require a method to fit object models to a 3D environment representation. In this work, we use the Iterative Closest Points (ICP) algorithm \cite{besl1992method}. ICP calculates the transformation that optimally aligns two sets of corresponding points: 
\begin{equation}
    \argmin_{\textbf{R},\textbf{t}}\sum\limits_{i} w_{i}\left(\textbf{R}\textbf{p}_{i}+\textbf{t}-\textbf{q}_{i}\right)
\end{equation}
where $\textbf{R}$ and $\textbf{t}$ are the optimal rotation and translation, $i$ is the point index, $w$ is a weighting factor, $\textbf{p}$ is the source point (e.g., mesh), and $\textbf{q}$ is the target point (e.g., scene). After the transformation is applied, the algorithm iteratively calculates new correspondences and new transformations until convergence. This method can be applied to meshes in a 3D point cloud scene by sampling the mesh. The optimization can be solved in closed form and computed using the Singular Value Decomposition (SVD) \cite{eggert1997estimating}. To provide improved robustness to occlusions and partial views, we reject outlier samples where the distance between corresponding points in the two sets, $d_{i}$, exceeds the model diameter and also weigh the correspondences based on the distance.
\begin{equation}
    w_{i} = (1+d_{i})^{-1}
\end{equation}
The ICP algorithm often converges to local minima because of the nonlinearity introduced by iteratively identifying mesh-scene correspondences. To reduce the likelihood of local minima, we use a random restarts procedure, where the initial conditions are drawn from random orientations and translation offsets (bounded by the model diameter) from the initial user-selected point.

\subsection{Interaction Flow}
Our method was designed to minimize the amount of input required from the operator. The operator provides corrections via a remote system which consists of a visualization of the environment (e.g., monitor, VR) and an input for providing corrections (e.g., 6D input, VR controller). Objects are registered in the environment as follows:

\begin{itemize}
  \item The user selects a point in the scene which represents an object of interest. This initialization circumvents the need for scene segmentation.
  \item All candidate models are fit using the Random Restarts ICP approach described in \S\ref{sec:tech}. For each model, the  inverse of the weighted residual is used as a likelihood.
  \item The mostly-likely fit is displayed to the user. If the model is not correct, the user can cycle to the next most likely fit. If the pose is not correct, the user provides relative corrective input. After the input, the system refits the object. This process of providing corrections and system refitting can be performed iteratively.
\end{itemize}

\subsection{Implementation}
We have developed an initial implementation in the Python programming language. The implementation is centered around a \emph{AffordanceEngine} class that keeps track of all objects of interest, models, and fits. The implementation also contains a ROS wrapper which allows for visualization in RViz \cite{kam2015rviz} and corrections using methods such as a SpaceMouse\textregistered\footnote{3Dconnexion. \emph{Trade names and trademarks are used in this report for identification only. Their usage does not constitute an official endorsement, either expressed or implied, by the National Aeronautics and Space Administration.}} and RViz interactive markers \cite{gossow2011interactive}. When using a relative input method (e.g., SpaceMouse), the inputs are mapped to be consistent with the current camera view orientation. In the future, we plan to implement our method and further functionality in C++ to improve performance.

\begin{figure}[t]
\centering
\includegraphics[width=3.30in]{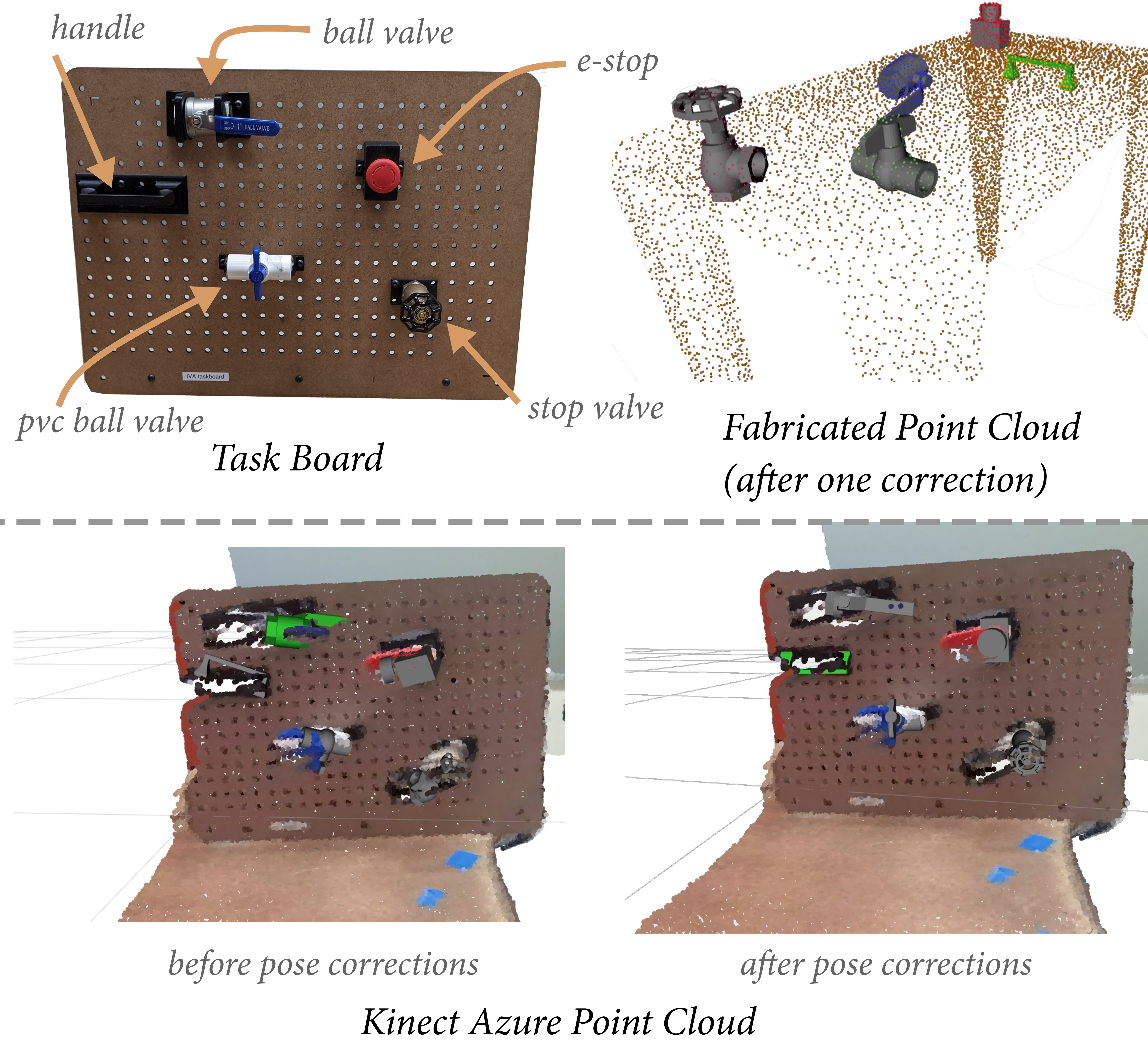}
\vspace{-5pt}
\caption{\emph{Top}: taskboard and corrected fits on the fabricated point cloud. \emph{Bottom}: before and after corrections on the Kinect point cloud.}
\label{fig:results}
\vspace{-10pt}
\end{figure}

\subsection{Proof-of-Concept Results}
We constructed a taskboard consisting of five prototypical objects (three valves, one e-stop, and one handle) encountered during ISS intra-vehicular activities (IVAs). We conducted a test using the SpaceMouse for corrections, a fabricated point cloud of the five meshes, and a point cloud of the taskboard captured using a Kinect Azure RGB-D camera. In the fabricated scene, the system was able to fit all five meshes. All models were correctly identified and only one mesh, the ball valve, required a correction to orientation. The kinect point cloud contained significant image distortion which complicated the automated fitting. In the Kinect point cloud, the system was able to fit all objects except the handle, which was not properly imaged from the camera. The mean fitting time was 4.84 seconds. The models required an average of 2.4 model changes (range 0-4). Each fit required corrections to the object pose.
\section{DISCUSSION AND FUTURE WORK}
Our preliminary results indicate the feasibility of our corrective approach to decrease user input during AT registration. However, the results from the Kinect highlight the criticality of the point cloud quality. For poorly captured views, the system will be unable to fit. In such cases, it will be important to add additional user options, such as user overrides or viewpoint adaptation.

Our current method contains a number of limitations which we plan to address in ongoing work. Our method assumes object models can be fit from a static mesh model, which is often not the case for objects that articulate. We plan to extend the fitting and corrections to include object articulations, such as the rotation angle of a valve. Our method relies on known objects specified in a library. To allow generalization to a wider range of objects, we plan to explore scale invariance (i.e., objects with the same geometry with different sizes) and semi-automated corrections to affordance primitive models \cite{pettinger2020reducing}. Finally, we plan to quantify the impact of the approach through both a comparative user study to existing methods and through situated demonstrations using the NASA Valkyrie robot platform \cite{radford2015valkyrie}.


\bibliographystyle{IEEEtran}
\bibliography{references}

\end{document}